\documentclass{article}

\usepackage{color}
\usepackage{amssymb}
\usepackage[cmex10]{amsmath}

\newtheorem{lemma}{Lemma}
\newtheorem{remark}{Remark}
\usepackage{spconf,amsmath,epsfig}
\usepackage{algorithm,algorithmic}
\usepackage{threeparttable}
\usepackage{tabularx,booktabs,graphicx,epsfig,subfigure}
\usepackage{float}
\usepackage{multirow}

\begin{document}\sloppy

\def\x{{\mathbf x}}
\def\L{{\cal L}}

\title{Graph Regularized Tensor Sparse Coding for Image Representation}
%
\name{Fei Jiang$^{1}$, Xiao-Yang Liu$^{1,2}$, Hongtao Lu$^{1}$, Ruimin Shen$^{1}$}
\address{$^1$Department of Computer Science and Engineering, Shanghai Jiao Tong University\\
         $^2$Department of Electrical Engineering, Columbia University}


\maketitle

\begin{abstract}
Sparse coding (SC) is an unsupervised learning scheme that has received an increasing amount of interests in recent years. However, conventional SC vectorizes the input images, which destructs the intrinsic spatial structures of the images. In this paper, we propose a novel graph regularized tensor sparse coding (GTSC) for image representation. GTSC preserves the local proximity of elementary structures in the image by adopting the newly proposed tubal-tensor representation. Simultaneously, it considers the intrinsic geometric properties by imposing graph regularization that has been successfully applied to uncover the geometric distribution for the image data. Moreover, the returned sparse representations by GTSC have better physical explanations as the key operation (i.e., circular convolution) in the tubal-tensor model preserves the shifting invariance property. Experimental results on image clustering demonstrate the effectiveness of the proposed scheme.
\end{abstract}
\section{Introduction}
\label{sec:intro}
Sparse coding (SC), which encode the images using only a few active coefficients, has been successfully applied to many areas across computer vision and pattern recognition \cite{zhang2015survey,gangeh2015supervised,wright2010sparse}, since it is computationally efficient and has physical interpretations.

However, conventional SC \cite{lee2006efficient} for image representations suffers from the following two major problems: (i) the vectorization preprocess breaks apart the local proximity of pixels and destructs the object structures of images; and (ii) the geometric distributions of the image space are ignored, while such information can significantly enhance the learning performance.

Two different kinds of sparse coding models have been proposed to preserve the intrinsic spatial structures of images: tensor sparse coding (TenSR) \cite{qi2016tensr,qi2013two} and convolutional sparse coding (CSC) \cite{bristow2013fast,heide2015fast}. For TenSR models \cite{qi2016tensr,qi2013two}, tensors are exploited for the image representation and a series of separable dictionaries are used to approximate the structures in each mode of images. Though the spatial structures are preserved by tensor representations, the relationships between the learned sparse coefficients and dictionaries are more complicated, which will cause the encoding (sparse coefficients) hard to interpret. For CSC models \cite{bristow2013fast,heide2015fast}, images are represented as the summation of the convolutions of  the filters that capture the local patterns of images and the corresponding feature maps. Each feature map has nearly the same size as the image, which significantly increases the computational complexity for further analyses of images, such as image classification and clustering.  And Figure \ref{fig1} shows the fundamental theoretical differences of the above-mentioned sparse coding models. Moreover, those  two kinds of models do not consider the geometric structures of the image space.

\begin{figure}[!t]
\centering
\includegraphics[height=0.17\textwidth]{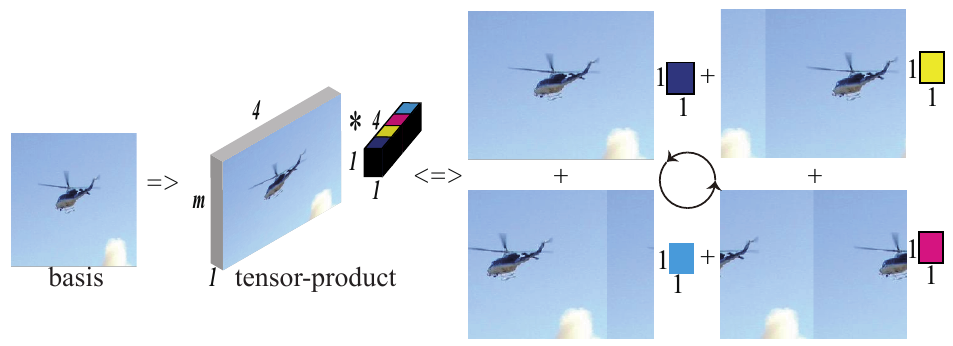}
\caption{Shifted versions of basis based on the tensor-product. The shifted versions correspond to a dynamic flight in a counter-clockwise direction.\label{fig2}}
\end{figure}

\begin{figure*}[!t]
\centering
\includegraphics[height=0.35\textwidth]{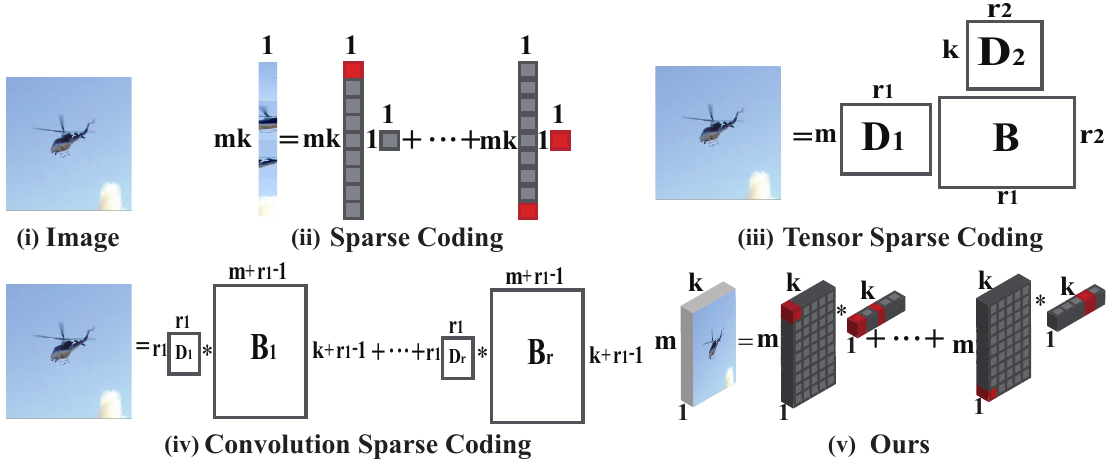}
\caption{Four sparse coding models. (i) the original image of size $m\times k$; (ii) conventional sparse coding; (iii) tensor sparse coding based on the tucker decomposition where $\mathbf{X}=\mathbf{B}\times_1\mathbf{D}_1\times_2\mathbf{D}_2$; (iv) convolution sparse coding (CSC) based on convolution operation where $\mathbf{X}=\sum_{j=1}^r\mathbf{D}_j\ast\mathbf{B}_j$. The size of $\mathbf{B}_j$ is almost the same as that of $\mathbf{X}$; (v) our tubal-tensor model based on circular the convolution operation where $\mathcal{X}=\sum_{j=1}^r\mathcal{D}_j\ast\mathcal{B}_j$. The size of $\mathcal{B}_j$ is much smaller than that in CSC.\label{fig1}}
\end{figure*}

Several sparse coding models incorporating the geometrical structures of the images space have been proposed. They are based on the locally invariant idea, which assumes that two close points in the original space are likely to have similar encodings. It has been shown \cite{zheng2011graph} that the learning performance can be significantly enhanced if the geometrical structure is exploited and the local invariance is considered. However, these sparse coding models ignore the spatial structure of the images due to the vectorization preprocess.

Motivated by the progress in tubal-tensor representation \cite{kilmer2013third,yang}, in this paper, we propose a novel graph regularized tensor sparse coding (GTSC) scheme for image representation that simultaneously considers the spatial structures of images and geometrical distributions of the image space. Firstly, we propose a novel tensor sparse coding model based on the tensor-product operation, which preserves the spatial structures of images by tensor representation. Unlike TenSR \cite{qi2016tensr,qi2013two}, the learned coefficients by our model have better physical explanations, which show the contributions of corresponding bases and their shifted versions, as shown in Figure \ref{fig2}. Then we incorporate the geometric distributions of the image space by  using the graph Laplacian as a smooth regularizer to preserve the local geometrical structures. By perserving the locally invariant property, GTSC has better discriminating power than the conventional SC \cite{lee2006efficient}.

The rest of this paper is organized as follows: Section \ref{sec:relat} introduces the proposed tubal-tensor sparse representation of images. Section \ref{sec:GTSC} presents the proposed GTSC model and the alternating minimization algorithm for GTSC. The experimental results on image clustering are presented in Section \ref{sec:eval}. Finally, we conclude the paper in Section \ref{sec:conclud}.
\section{Tubal-tensor Sparse Representation}
\label{sec:relat}
\subsection{Notation}
A third-order tensor of size $m\times n\times k$ is denoted as $\mathcal{X}\in\mathbb{R}^{m\times n\times k}$. $\mathcal{X}^{(\ell)}$ represents the $\ell$-th frontal slice $\mathcal{X}(:,:,\ell)$ which is a matrix, $\underline{\mathcal{X}}\in\mathbb{R}^{mk\times nk}$ represents the expansion of $\mathcal{X}$ along the third dimension where $\underline{\mathcal{X}}=[\mathcal{X}^{(1)};\mathcal{X}^{(2)};\cdots;\mathcal{X}^{(k)}]$,  $\widehat{\mathcal{X}}$ is the discrete Fourier transform (DFT) along the third dimension of $\mathcal{X}$, and $\mathcal{X}^{\dagger}$ represents the transpose of $\mathcal{X}$ where $\mathcal{X}^{\dagger^{(1)}}=\mathcal{X}^{(1)^T}$ and $\mathcal{X}^{\dagger^{(\ell)}}=\mathcal{X}^{(k+1-\ell)^T}$, $2\leq \ell\leq k$. The superscript ``$T$" denotes the transpose of matrices.

For convenience, tensor spaces $\mathbb{R}^{1\times1\times k}$, $\mathbb{R}^{m\times1\times k}$, $\mathbb{R}^{m\times n\times k}$ are denoted as $\mathbb{K}$, $\mathbb{K}^m$, and $\mathbb{K}^{m\times n}$, respectively. $[k]$ denotes the set $\{1,2,\cdots,k\}$. The $\ell_1$ and Frobenius norms of tensors are denoted as $\|\mathcal{X}\|_1=\sum_{i,j,k}|\mathcal{X}(i,j,k)|$, and $\|\mathcal{X}\|_F^2=\left(\sum_{i,j,k}\mathcal{X}(i,j,k)^2\right)^{1/2}$, respectively.

\subsection{Tensor-linear Combination}
A two-dimensional image of size $m\times k$ is represented by a third-order tensor $\mathcal{X}\in\mathbb{K}^{m}$, which can be approximated by the tensor-product  between $\mathcal{D}\in\mathbb{K}^{m\times r}$ and $\mathcal{B}\in\mathbb{K}^r$ as
\begin{equation}
\label{eq:t-prod}
\mathcal{X} = \mathcal{D} * \mathcal{B},
\end{equation}
where $*$ denotes tensor-product introduced in \cite{kilmer2013third}.

Note that, $\mathcal{X}$ can be rewritten as a tensor-linear combination of tensor bases $\{D(:,j,:)\}_{j=1}^r\subset\mathbb{K}^m$ with corresponding tensor representations $\{\mathcal{B}(j,1,:)\}_{j=1}^r\subset\mathbb{K}$:
\begin{equation}
\label{eq:t-lin}
\begin{split}
 \mathcal{X}&=\mathcal{D}*\mathcal{B}\\
            &=\mathcal{D}(:,1,:)*\mathcal{B}(1,1,:)+\cdots+\mathcal{D}(:,r,:)*\mathcal{B}(r,1,:).
\end{split}
\end{equation}
Equation (\ref{eq:t-lin}) is quite similar to linear combinations as tubes $\mathcal{B}(r,1,:)$ play the same role as scalars in the matrix representation.

\subsection{Tubal-tensor Sparse Representation}

Given $n$ images of size $m\times k$, we present them as  a third-order tensor $\mathcal{X}\in\mathbb{K}^{m\times n}$. Let $\mathcal{D}\in\mathbb{K}^{m\times r}$ be the tensor dictionary, where each lateral slice $\mathcal{D}(:,j,:)$ represents a tensor basis, and $\mathcal{S}\in\mathbb{K}^{r\times n}$ be the tensor corresponding representations. Each image $\mathcal{X}(:,j,:)$ is approximated by a sparse tensor-linear combination of those tensor bases. Our tubal-tensor sparse coding (TubSC) model can be formulated as:
\begin{eqnarray}
\label{obj:tsc}
 \min_{\mathcal{D},\mathcal{B}} && \frac{1}{2}\|\mathcal{X} - \mathcal{D} * \mathcal{B}\|_F^2 + \beta\|\mathcal{B}\|_{1}\nonumber\\
 \rm{s.t.} && \|\mathcal{D}(:,j,:)\|_F^2 \leq 1,\quad j\in[r],
\end{eqnarray}
where $\beta$ is the sparsity regularizer.
\begin{remark}
Conventional SC is a special case of TubSC.
\end{remark}

\subsection{Explanations of Tensor Representations}
\label{sec:exp}
To explain the tensor representations, we introduce Lemma \ref{lem1} which bridges the tensor-product with the matrix-product.
\begin{lemma}\cite{kilmer2013third}
\label{lem1}
The tensor-product $\mathcal{X}=\mathcal{D}*\mathcal{B}$ has an equivalent  matrix-product as:
\begin{equation}
\label{eq:circ}
\underline{\mathcal{X}} = \underline{\mathcal{D}}^c\underline{\mathcal{B}},
\end{equation}
where $\underline{\mathcal{D}}^c$ is the circular matrix of $\mathcal{D}$ defined as follows:
\begin{equation}
\label{eq:cirmat}
 {\underline{D}}^c = \left(
               \begin{array}{cccc}
                 \mathcal{D}^{(1)} & \mathcal{D}^{(k)} & \cdots & \mathcal{D}^{(2)} \\
                 \mathcal{D}^{(2)} & \mathcal{D}^{(1)} & \cdots & \cdots \\
                 \cdots & \cdots & \cdots & \mathcal{D}^{(k)} \\
                 \mathcal{D}^{(k)}& \mathcal{D}^{(k-1)} & \cdots &\mathcal{D}^{(1)}\\
               \end{array}
             \right).
\end{equation}
\end{lemma}

Assuming the vectorization formulations of tensor bases are $\mathbf{D}=[d_1,\cdots,d_r]\in\mathbb{R}^{mk\times r}$, where $d_j=\mathcal{D}(:,j,:)(:)$, $j\in[r]$, then $\underline{\mathcal{D}}^c(:,1:r)$ is actually $\mathbf{D}$. Moreover, $\underline{\mathcal{D}}^c$ are the set of shifted versions of $\mathbf{D}$, which can be denoted as $\underline{\mathcal{D}}^c=[\mathbf{D},\mathbf{D}_1,\cdots,\mathbf{D}_{k-1}]$.

If we rewrite $\underline{\mathcal{B}}=[b;b_1;\cdots;b_{k-1}]\in\mathbb{R}^{rk}$ with corresponding to the shifted versions of $\mathbf{D}$, the tensor-product (\ref{eq:t-prod}) is  further transformed into linear combination as follows:
\begin{equation}
\label{eq:t-lin-eqv}
\underline{\mathcal{X}} = \underline{\mathcal{D}}^c\underline{\mathcal{B}}=\mathbf{D}b+\mathbf{D}_1b_1+\cdots+\mathbf{D}_{k-1}b_{k-1}.
\end{equation}
From (\ref{eq:t-lin-eqv}), we can see the explicit meanings of tensor representations $\mathcal{B}$, which display the reconstruction contributions of the  corresponding original bases and the shifted versions of bases simultaneously. Figure \ref{fig2} shows the shifted versions of an image basis. It can be seen that the shifted versions of the basis are used for image reconstruction without storing them.

\section{Graph Regularized Tensor Sparse Coding}
\label{sec:GTSC}
In this section, we present our graph regularized tensor sparse coding (GTSC) model which simultaneously takes into account the spatial structures of images and local geometric information of the image space.

\subsection{Problem Formulation}
From one aspect, TubSC model defined by (\ref{obj:tsc}) can preserve the spatial structures of images based on the tensor-linear combinations in (\ref{eq:t-lin}). From another aspect, one might further hope that the learned tensor dictionary can respect the intrinsic geometrical information of the image space. A natural assumption is to keep local invariance where  the learned sparse representations of two close points in the original space are also close to each other. This assumption is usually referred to as manifold learning \cite{zheng2011graph}, which can significantly enhance the learning performance.

Given a set of $n$ images $\{\mathbf{X}_j\}_{j=1}^n$ of size $m\times k$, a $q$-nearest neighbor graph  $\mathbf{W}\in\mathbb{R}^n$ is constructed.
Considering the problem of mapping the weighted graph to the tensor sparse representation $\mathcal{B}$, we first make an expansion of $\mathcal{B}$ along the third dimension as $\underline{\mathcal{B}}\in\mathbb{R}^{mk\times n}$ , where each column is a sparse representation of an image. A reasonable criterion for choosing a better mapping is to minimize the following objective function:
\begin{equation}
\label{def:reg}
\frac{1}{2}\sum_{ij}\|\underline{\mathcal{B}}(:,i)-\underline{\mathcal{B}}(:,j)\|_F^2\mathbf{W}_{ij}=\text{Tr}(\underline{\mathcal{B}}\mathbf{L}\underline{\mathcal{B}}^T),
\end{equation}
where $\text{Tr}(\cdot)$ represents the trace of the matrix, and $\mathbf{L}=\mathbf{E}-\mathbf{W}$ is the Laplacian matrix, where $\mathbf{E}=\text{diag}(e_1,\cdots,e_n)$, and $e_i=\sum_{j=1}^{n}\mathbf{W}_{ij}$.

By incorporating the Laplacian regularizer (\ref{def:reg}) into the TubSC model (\ref{obj:tsc}), we propose a novel model named graph regularized tensor sparse coding (GTSC) as:
\begin{eqnarray}
\label{obj:gtsc}
 \min_{\mathcal{D},\mathcal{B}} && \frac{1}{2}\|\mathcal{X} - \mathcal{D} * \mathcal{B}\|_F^2 +\alpha\text{Tr}(\underline{\mathcal{B}}\mathbf{L}\underline{\mathcal{B}}^T) + \beta\|\mathcal{B}\|_{1}\nonumber\\
 \rm{s.t.} && \|\mathcal{D}(:,j,:)\|_F^2 \leq 1,\quad j\in[r],
\end{eqnarray}
where $\alpha\geq0$ is the graph regularizer, and $\beta\geq0$ is the sparsity regularizer.

Problem (\ref{obj:gtsc}) is quite challenging due to the non-convex objective function and the convolutional operation. Instead of transforming (\ref{obj:gtsc}) into conventional graph regularized SC formulation based on Lemma \ref{lem1}, we propose a much more efficient algorithm by alternatively optimizing $\mathcal{D}$ and $\mathcal{B}$ directly in the tensor space.
\begin{algorithm}[t!]
\caption{Iterative Shrinkage Thresholding algorithm based on Tensor representation (ISTT) \label{alg1}}
\begin{algorithmic}[1]
\STATE{\textbf{Input:} $n$ images: $\mathcal{X}\in\mathbb{R}^{m\times n}$, dictionary: $\mathcal{D}$, regularizers: $\alpha\geq0$, $\beta\geq0$, graph Laplacian: $\mathbf{L}$, maximum iterative steps: num,}
\STATE{\textbf{Initialization:} Set $\mathcal{C}_1=\mathcal{B}_0\in\mathbb{K}^r$, $t_1=1$,}
\FOR {iter = 1 to \text{num} }
     \STATE{Set $\text{Lip}^{\text{iter}} = \eta^{\text{iter}}(\sum_{\ell=1}^{k}\|\widehat{\mathcal{D}}^{(\ell)^H}\widehat{\mathcal{D}}^{\ell}\|_F+2\alpha\|\mathbf{L}\|_2)$,}
     \STATE{Compute $\nabla f(\mathcal{C}_{\text{iter}})$ via Equation (\ref{eq:grad}),}
     \STATE{Compute $\mathcal{B}_{\text{iter}}$ via $\textbf{Prox}_{\beta/\text{Lip}^{\text{iter}}}(\mathcal{C}_k-\frac{1}{\text{Lip}^{\text{iter}}}\nabla f(\mathcal{C}_{\text{iter}}))$,}
     \STATE{$t_{\text{iter}+1}=\frac{1+\sqrt{1+4t^2_{\text{iter}}}}{2}$,}
     \STATE{$\mathcal{C}_{{\text{iter}}+1}=\mathcal{B}_{\text{iter}} + \frac{t_{\text{iter}}-1}{t_{\text{iter}+1}}(\mathcal{B}_{\text{iter}}-\mathcal{B}_{\text{iter}-1})$,}
\ENDFOR
\STATE{\textbf{Output:} Sparse coefficients $\mathcal{B}$.}
\end{algorithmic}
\end{algorithm}
\subsection{Graph Regularized Tensor Sparse Representations $\mathcal{B}$}
In this subsection, we discuss how to solve (\ref{obj:gtsc}) by fixing the tensor dictionary $\mathcal{D}$. Problem (\ref{obj:gtsc}) becomes:
\begin{equation}
\label{obj:b}
\min_{\mathcal{B}\in\mathbb{K}^r}\frac{1}{2}\|\mathcal{X}-\mathcal{D}*\mathcal{B}\|_F^2 + \alpha\text{Tr}(\underline{\mathcal{B}}\mathbf{L}\underline{\mathcal{B}}^T) + \beta\|\mathcal{B}\|_{1}
\end{equation}
By Lemma 1, (\ref{obj:b}) is equivalent to:
\begin{equation}
\label{obj:b1}
\min_{\underline{S}\in\mathbb{R}^{rk}}\frac{1}{2}\|\underline{\mathcal{X}}-\underline{\mathcal{D}}^c\underline{\mathcal{B}}\|_F^2 + \alpha\text{Tr}(\underline{\mathcal{B}}\mathbf{L}\underline{\mathcal{B}}^T) + \beta\|\underline{\mathcal{B}}\|_{1}.
\end{equation}
The size of the dictionary $\underline{\mathcal{D}}^c\in\mathbb{R}^{mk\times rk}$ in (\ref{obj:b1}) will be significantly increased for high dimensional images, which will need more computational resources.

To alleviate the above-mentioned problem, we propose a novel Iterative Shrinkage Thresholding algorithm based on the tensor presentation (ISTT) to solve (\ref{obj:b}) directly, which is rewritten as:
\begin{equation}
\label{obj:b2}
\min_{\mathcal{B}}f(\mathcal{B})+\beta g(\mathcal{B}),
\end{equation}
where $f(\mathcal{B})$ stands for $\frac{1}{2}\|\mathcal{X}-\mathcal{D}*\mathcal{B}\|_F^2 + \alpha\text{Tr}(\underline{\mathcal{B}}\mathbf{L}\underline{\mathcal{B}}^T)$, and $g(\mathcal{B})$ stands for the sparsity constraint $\|\mathcal{B}\|_{1}$.

Then, the iterative shrinkage function is constructed by the linearized function around the previous estimation of $\mathcal{B}_{p}$ with the proximal regularization and the nonsmooth regularization. Thus, at the $p+1$-th iteration, $\mathcal{B}_{p+1}$ is updated by:
\begin{equation}
\label{obj:b3}
\begin{aligned}
\mathcal{B}_{p+1} = \arg\min_{\mathcal{B}}&f(\mathcal{B}_{p}) + \langle\nabla f(\mathcal{B}_{p}),\mathcal{B}-\mathcal{B}_{p}\rangle\\
&+ \frac{L_{p+1}}{2}\|\mathcal{B}-\mathcal{B}_{p}\|_F^2 + \beta g(\mathcal{B}),
\end{aligned}
\end{equation}
To solve (\ref{obj:b3}), we firstly show $\nabla f(\mathcal{B})$ w.r.t. the data reconstruction term $\frac{1}{2}\|\mathcal{X}-\mathcal{D}*\mathcal{B}\|_F^2$:
\begin{equation}
\label{eq:grad}
\nabla f(\mathcal{B}) = \mathcal{D}^{\dagger}*\mathcal{D}*\mathcal{B} - \mathcal{D}^{\dagger}*\mathcal{X} + 2\alpha\mathcal{B}*\mathcal{L},
\end{equation}
where $\mathcal{L}\in\mathbb{K}^{n\times n}$ with the first frontal slice $\mathcal{L}(:,:,1)=\mathbf{L}$ and the other slices $\mathcal{L}(:,:,\ell)=0$, $\ell\in\{2,\cdots,k\}$.


Secondly, we discuss how to determine the Lipschitz constant $\text{Lip}_{p+1}$ in (\ref{obj:b3}). For every $\mathcal{B}$ and $\mathcal{C}$, we have
\begin{equation}
\label{eq:L}
\begin{aligned}
&\|\nabla f(\mathcal{B}) - \nabla f(\mathcal{C})\|_F\\
&=\|\mathcal{D}^{\dagger}*\mathcal{D}*(\mathcal{B}-\mathcal{C})+2\alpha(\mathcal{B}-\mathcal{C})*\mathcal{L}\|_F\\
&=\sum_{j=1}^n\|\underline{(\mathcal{D}^{\dagger}*\mathcal{D})}^c(\underline{\mathcal{B}_j}-\underline{\mathcal{C}_j})+2\alpha(\underline{\mathcal{B}_j}-\underline{\mathcal{C}_j})\mathbf{L}\|_2\\
&\leq(\|\underline{(\mathcal{D}^{\dagger}*\mathcal{D})}^c\|_2+2\alpha\|\mathbf{L}\|_2)\sum_{j=1}^n\|\underline{\mathcal{B}_j}-\underline{\mathcal{C}_j}\|_2\\
&\leq(\sum_{\ell=1}^k\|\widehat{\mathcal{D}}^{(\ell)^H}\widehat{\mathcal{D}}^{(\ell)}\|_F^2+2\alpha\|\mathbf{L}\|_2)\|\mathcal{B} - \mathcal{C}\|_F,
\end{aligned}
\end{equation}
where the superscript ``$H$" represents conjugate transpose, and $\mathcal{B}_j=\mathcal{B}(:,j,:)$.

Thus the Lipschitz constant of $f(\mathcal{B})$ used in our algorithm is $\text{Lip}(f)=\sum_{\ell=1}^k\|\hat{\mathcal{D}}^{(\ell)^H}\hat{\mathcal{D}}^{(\ell)}\|_F^2+2\alpha\|\mathbf{L}\|_2$.

Lastly, (\ref{obj:b2}) can be solved by the proximal operator $\textbf{Prox}_{\beta/L_{p+1}}(\mathcal{B}_{p}-\frac{1}{L_{p+1}}\nabla f(\mathcal{B}_{p}))$, where $\textbf{Prox}_{\tau}$ is the soft-thresholding operator $\textbf{Prox}_{\tau}(\cdot)\to \text{sign}(\cdot)\max(|\cdot| - \tau,0)$.

To speed up the convergence of the iterative shrinkage algorithm, an extrapolation operator is adopted \cite{qi2016tensr}. Algorithm \ref{alg1} summarizes our proposed Iterative Shrinkage Thresholding algorithm based on tensor presentation (ISTT).
\begin{algorithm}[t!]
\caption{Algorithm for GTSC \label{alg2}}
\begin{algorithmic}[1]
 \renewcommand{\algorithmicrequire}{\textbf{Input:}}
 \renewcommand{\algorithmicensure}{\textbf{Output:}}
 \REQUIRE $n$ images: $\mathcal{X}\in\mathbb{K}^{m\times n}$, the number of atoms: $r$, regularizers: $\alpha\geq0$, $\beta\geq0$, graph Laplacian $\mathbf{L}$, maximum iterative steps: \text{num},
 \noindent\STATE{\textbf{Initialization:} randomly initialize $\mathcal{D}\in\mathbb{K}^{m\times r}$, $\mathcal{B}:=0\in\mathbb{K}^{r\times n}$, and Lagrange dual variables $\lambda\in\mathbb{R}^r$,}
 \FOR{$\text{iter}=1$ to \text{num}}
   \STATE{//\textbf{Graph Regularized Tensor Sparse Representation}}\\
   \STATE{Solving $\mathcal{B}$ via Equation (\ref{obj:b3}) in Algorithm \ref{alg1},}
   \STATE{//\textbf{Tensor Dictionary Learning}}\\
   \STATE{$\widehat{\mathcal{X}}=\text{fft}(\mathcal{X},[~],3)$, $\widehat{\mathcal{B}}=\text{fft}(\mathcal{B},[~],3)$,}
   \FOR{$\ell=1$ to $k$}
    \STATE{Solving (\ref{dual}) for $\Lambda$ by Newton's method,}
    \STATE{Calculate $\widehat{\mathcal{D}}^{(\ell)}$ from (\ref{dic}),}
   \ENDFOR
   \STATE{$\mathcal{D}=\text{ifft}(\mathcal{D},[~],3)$,}
 \ENDFOR
\ENSURE $\mathcal{D}$, $\mathcal{B}$.
\end{algorithmic}
\end{algorithm}
\subsection{Tensor Dictionary Learning $\mathcal{D}$}
For learning the dictionary $\mathcal{B}$ while fixed $\mathcal{S}$, the optimization problem is:
\begin{eqnarray}
\label{2DSC_D}
\min_{\mathcal{D}} && \frac{1}{2}\|\mathcal{X}-\mathcal{D}*\mathcal{B}\|_F^2\nonumber\\
\rm{s.t.} && \|\mathcal{D}(:,j,:)\|_F^2\leq 1, j\in[r],
\end{eqnarray}
where atoms are coupled together due to the circular convolution operation. Therefore, we firstly decompose (\ref{2DSC_D}) into $k$ nearly-independent problems (that are coupled only through the norm constraint) by DFT as follows:
\begin{eqnarray}
\label{basis2}
 \min_{\widehat{\mathcal{D}}^{(\ell)},\ell\in[k]} && \sum_{\ell=1}^{k}\|\widehat{\mathcal{X}}^{(\ell)}-\widehat{\mathcal{D}}^{(\ell)}\widehat{\mathcal{B}}^{(\ell)}\|_F^2\nonumber\\
 \rm{s.t}. && \sum_{\ell=1}^{k}\|\widehat{\mathcal{D}}^{(\ell)}(:,j)\|_F^2\leq k, j\in[r].
\end{eqnarray}

Then, we adopt the Lagrange dual \cite{lee2006efficient} for solving the dual variables by Newton's algorithm. Another advantage of Lagrange dual is that the number of optimization variables is $r$, which is much smaller than $mkr$ of the primal problem for solving $\mathcal{D}$.

To use the Lagrange dual algorithm, firstly, we consider the Lagrangian of (\ref{basis2}):
\begin{eqnarray}
 \text{Lag}(\widehat{\mathcal{D}},\Lambda)&=&\sum_{\ell=1}^k\|\widehat{\mathcal{X}}^{(\ell)}-\widehat{\mathcal{D}}^{(\ell)}\widehat{\mathcal{B}}^{(\ell)}\|_F^2+\nonumber\\
 &&\sum_{j=1}^r\lambda_j\left(\sum_{\ell=1}^k\|\widehat{\mathcal{D}}^{(\ell)}(:,j)\|_F^2-k\right),
\end{eqnarray}
where $\lambda_j\geq0$, $j\in[r]$ is a dual variable, and $\Lambda=\text{diag}(\lambda)$.

Secondly, minimizing over $\widehat{\mathcal{D}}$ analytically, we obtain the optimal formulation of $\widehat{\mathcal{D}}$:

\begin{equation}
\label{dic}
\widehat{\mathcal{D}}^{(\ell)} = \left(\widehat{\mathcal{X}}^{(\ell)}\widehat{\mathcal{B}}^{(\ell)^H}\right)\left(\widehat{\mathcal{B}}^{(\ell)}\widehat{\mathcal{B}}^{(\ell)^H}+\Lambda\right)^{-1}, \ell\in[k].
\end{equation}
Substituting this expression into the Lagrangian $\mathcal{L}(\widehat{\mathcal{D}},\Lambda)$, we obtain the Lagrange dual function $\mathcal{D}(\Lambda)$, and the optimal dual variables by using Newton's method.
\begin{equation}
\label{dual}
\mathcal{D}(\Lambda)=-\sum_{\ell=1}^k\text{Tr}\left(\widehat{\mathcal{D}}^{(\ell) ^H}\widehat{\mathcal{X}}^{(\ell)}\widehat{\mathcal{B}}^{(\ell)^H}\right)-k\sum_{j=1}^r\lambda_j.
\end{equation}
Once getting the dual variables, the dictionary can be recovered using Equation (\ref{dic}).

The algorithm we proposed for GTSC is shown in Algorithm \ref{alg2}

\subsection{Complexity Analysis}
Given $n$ images of size $m\times k$, the numbers of bases $r$ and $q$ nearest neighbors, the computational complexities of GTSC and GraphSC \cite{zheng2011graph} are as follows:

For sparse representation learning, GTSC is based on an iterative shrinkage thresholding algorithm in the tensor space, and GraphSC is based on a feature-sign algorithm. The computational complexity for GTSC is $O(rmkn+klogk)$ and for GraphSC\cite{zheng2011graph} is $O(r0^3 + r0^2mkn)$, where $r0$ is the number of non-zero coefficients.

For dictionary learning, both GTSC and GraphSC\cite{zheng2011graph} are based on Lagrange-dual algorithms, and the computational complexities are $O(r^2n)$. For GTSC, the optimal dictionary is obtained slice by slice, and the computational complexity for each slice is $O(r^3+r^2n+rmn)$. For GraphSC, the computational complexity is $O(r^3+r^2n+rmnk)$.

Overall, the computational complexity of GTSC is less than GraphSC\cite{zheng2011graph}, especially for high dimensional data.

\section{Evaluation}
\label{sec:eval}
We apply our proposed algorithms, TubSC in (\ref{obj:tsc}) and GTSC in (\ref{obj:b1}) models, to image clustering tasks on four image databases: COIL20\footnote[1]{$http://www1.cs.columbia.edu/CAVE/software/softlib/coil-100.php$}, USPS\footnote[2]{$http://www.cad.zju.edu.cn/home/dengcai/Data/MLData.html$}, ORL\footnote[3]{$http://www.uk.research.att.com/facedatabase.html$}, Yale\footnote[4]{$http://www.cad.zju.edu.cn/home/dengcai/Data/FaceData.html$}. The important statistics of these datasets are sumarized in Table \ref{tab1}. Two metrics, the  accuracy (ACC) and the normalized mutual information (NMI), are used for evaluations. ACC measures the percentage of correct labels obtained by an algorithm and NMI measures how similar two clusters are. The details of these two metrics can refer to \cite{zheng2011graph}.
\begin{table}[t]
\begin{center}
\caption{Statistics of the Four Dataset} \label{tab1}
\small{\begin{tabular}{|c|c|c|c|}
  \hline
  Data & Class & Size & Number of Images\\
  \hline
  COIL20 & 20 & 32$\times$32 &1440 \\
  USPS    & 10  & 16$\times$16 &9298 \\
  ORL   & 40  & 32$\times$32 &400\\
  YALE & 15  & 64$\times$64 &165\\
  \hline
\end{tabular}}
\end{center}
\end{table}

\subsection{Compared Algorithms}

To evaluate the clustering performances, we compare our proposed methods against the conventional SC \cite{lee2006efficient} and GraphSC \cite{zheng2011graph}. The performance scores are obtained by averaging over the 10 tests. For each test, we first apply the compared methods to learn new representations for images, and then apply K-means in the new representation space. For SC \cite{lee2006efficient} and GraphSC \cite{zheng2011graph} methods, PCA is used to reduce the data dimensionality by retaining $98\%$ of the variance. The numbers of bases for USPS and YALE are set to 128, and those for COIL20 and ORL are set to 256, respectively. For our methods, we do not need to reduce the data dimensionality. Moreover, the numbers of bases are much smaller than those used in SC \cite{lee2006efficient} and GraphSC \cite{zheng2011graph}, due to the powerful representation generated from the tensor-product. For our methods, the numbers of bases are set to 45 for all data sets except YALE, which is set to 80. Based on the physical explanations of tensor sparse representations, we use C as the final image representation, which is defined as
\begin{equation}
\mathbf{C}(i,j) = \sqrt{\sum_{\ell}^k\mathcal{S}(i,j,\ell)^2}.
\end{equation}
For GraphSC\cite{zheng2011graph} and GTSC, we empirically set the graph regularization parameter alpha to 1 and the number of nearest neighbors to 3.

\subsection{Clustering Results}
Table \ref{tab2} shows the clustering results in terms of ACC and NMI. As can be seen, our GTSC algorithm performs the best in all the cases. TubSC performs much better than conventional SC, which indicate that by considering spatial proximity information of images, the learning performance can be significantly enhanced. Moreover, GraphSC outperforms SC, which shows that by encoding geometrical distribution information of the image space, the learning performance can also be improved.

We would like to point out that we use much smaller sizes of dictionaries in our proposed models than SC and GraphSC, but without any dimensionality reduction preprocessing.
For another aspect, we do not compare the clustering performances with TenSR \cite{qi2016tensr} and CSC \cite{heide2015fast}, which also consider the spatial structures of images. Without dimensionality reduction, the representations learned from TenSR and CSC are larger than the original images, which significantly increase the computational complexity of clustering.

\begin{table}[t]
\begin{center}
\small{\caption{Clustering Performances of Different Algorithms on Four Datasets} \label{tab2}
\begin{tabular}{|c|c|c|c|c|c|}
  \hline
  Data                         & COIL20 & USPS & ORL & YALE&AVG. \\
  \hline
          &\multicolumn{5}{|c|}{ACC ($\%$)} \\
  \hline
  K-Means                      &60.49   &67.45 &53.50& 51.52  &58.24   \\
  SC \cite{lee2006efficient}   &67.43   &68.62 &53.25& 54.55  &60.96 \\
  GraphSC \cite{zheng2011graph}&75.28   &67.89 &59.50& 57.58  &65.06 \\
  TubSC(Ours)                  &72.29   &71.16 & 60.00& 56.36 &64.95  \\
  GTSC(Ours)                   &\textbf{83.19} &\textbf{76.11}&\textbf{67.50}&\textbf{60.00} &\textbf{71.70}    \\
  \hline
          & \multicolumn{5}{|c|}{NMI ($\%$)}\\
  \hline
  K-Means &73.86   &62.20 &71.82& 53.69&65.39\\
  SC\cite{lee2006efficient}&73.24   &65.91 &72.18& 54.70&66.51 \\
  GraphSC \cite{zheng2011graph}&81.03   &67.22 &76.00& 60.21&71.12 \\
  TubSC(Ours)&80.52  &68.21   &76.94 &61.15 &71.70  \\
  GTSC(Ours) &\textbf{89.66} &\textbf{79.31}&\textbf{81.22}&\textbf{63.98}   &\textbf{78.54}\\
  \hline
\end{tabular}}
\end{center}
\end{table}
\section{Conclude}
\label{sec:conclud}
In this paper, we propose a novel graph regularized tensor sparse coding (GTSC) model for image presentation, which explicitly considers both the spatial proximity information of images and geometric structures of the image space. GTSC is based on a novel tubal-tensor sparse coding (TubSC) model where the tensor encodings of TubSC have richer explanations than conventional sparse coding. The experimental results on image clustering have demonstrated that our proposed algorithm can have better representation power and significantly enhance the clustering performance.

\section{Ackonwledgements}
This work is supported by NSFC (No.61671290) in China, the Key Program for International S$\&$T Cooperation Project (No.2016YFE0129500) of China  and partially supported by the Basic Research Project of Innovation Action Plan (No. 16JC1402800) of Shanghai Science and Technology Committee.
\bibliographystyle{IEEEbib}
\bibliography{GTSCref1}

\end{document}